\documentclass[letterpaper, 10 pt, conference]{ieeeconf}  %

\IEEEoverridecommandlockouts                              %

\usepackage{graphicx} %
\usepackage{xurl}     %
\usepackage{tabularx}
\usepackage{array}
\usepackage{pgfplots}
\pgfplotsset{compat=1.18}
\usepackage{tikz}
\usepackage{cite}
\usepackage{multirow}
\usepackage[export]{adjustbox}
\title{\LARGE \bf
Face versus Body Tracking for Human--Robot Interaction: \\An Egocentric Dataset and Evaluation
}

\author{Jessica Wenninger$^{1}$ and Gabriel Skantze$^{2}$%
\thanks{This project has received funding from the European Union's Horizon 2023 research and innovation programme under the Marie Sk\l{}odowska-Curie grant agreement No. 101168792 (SWEET).}%
\thanks{$^{1}$Jessica Wenninger is with Furhat Robotics, Stockholm, Sweden, and the University of Naples Federico II, Naples, Italy.
        {\tt\small jessi@furhatrobotics.com}}%
\thanks{$^{2}$Gabriel Skantze is with Furhat Robotics, Stockholm, Sweden, and the Division of Speech, Music and Hearing, KTH Royal Institute of Technology, Stockholm, Sweden.
        {\tt\small skantze@kth.se}}%
}

\begin{document}

\maketitle
\thispagestyle{empty}
\pagestyle{empty}

\begin{abstract}
Meaningful human-robot interaction (HRI) requires a robot to continuously assess user engagement through persistent user tracking. However, state-of-the-art Multi-Object Tracking models  are heavily optimized for surveillance or autonomous driving. A social robot faces distinct egocentric challenges, such as humans moving in unpredictable nonlinear patterns, obstructing each other, or leaving and reentering the scene. These dynamics trigger frequent identity switches (IDSW), causing the robot to lose its footing mid-conversation. To address this, we introduce a focused, custom-annotated egocentric dataset collected via the Furhat robot. We present a systematic evaluation isolating detection errors from tracking logic, comparing face versus body tracking, and assessing the impact of extended memory and appearance re-identification (ReID). Results indicate that increasing temporal memory mitigates prolonged occlusions but fails on complex dynamic events. Integrating ReID resolves complex switches but exhibits opposing effects: it substantially improves body tracking stability, yet causes facial IDSW to spike due to profile angle sensitivity. Ultimately, our optimized pipeline reduces IDSW by 49\% compared to a standard tracking-by-detection baseline, effectively mitigating interaction breakdowns. As standard benchmarks lack dense, close-quarter occlusions, this work highlights the critical need for natively captured social dynamics to truly validate HRI perception models.
\end{abstract}

\section{INTRODUCTION}

To enable meaningful human-robot interaction (HRI), a robot must be capable of continuously assessing and maintaining user engagement \cite{sorrentino_definition_2024}. This requires a robot to consistently track who it is interacting with over time \cite{del_duchetto_are_2020}. Crucially, this involves understanding the interaction's \textit{footing}: the ability to dynamically track and distinguish the specific roles of individuals in its environment, such as active interlocutors (addressees), bystanders, and overhearers (nonparticipants) \cite{mutlu_footing_2009}. Robust perception is the foundational building block for maintaining these roles during long-term interactions \cite{yang_online_2025}. If a tracking system suffers from frequent identity switches (IDSW)---where the algorithm incorrectly assigns a new ID to an existing person after an occlusion---the robot effectively loses this footing mid-conversation, degrading the user's perception of the robot's social competence \cite{tian_taxonomy_2021}.

\begin{figure}[t] %
    \centering
    \includegraphics[width=\columnwidth]{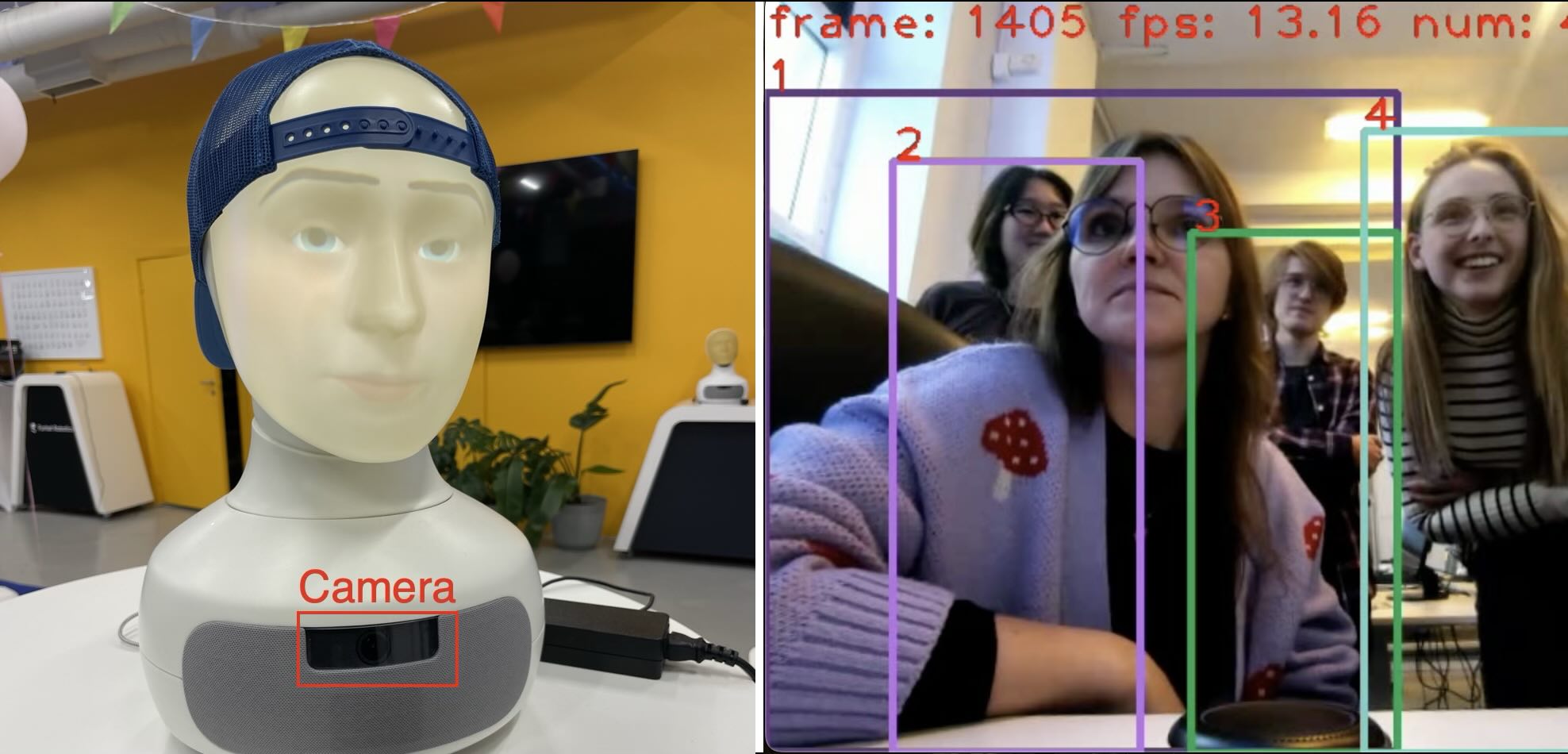}
    \caption{The Egocentric HRI Tracking Challenge. \textbf{Left:} The experimental setup with the Furhat robot in a real-world office environment. \textbf{Right:} A representative frame from the robot’s egocentric perspective. The scene highlights the difficulty of maintaining consistent identities for multiple actors despite dynamic background motion and severe occlusions.}
    \label{fig:system_overview}
\end{figure}

While modern object tracking models perform exceptionally well, there is a fundamental disconnect between computer vision and HRI. State-of-the-art models are heavily optimized for surveillance cameras \cite{taylor_regroup_2022} or autonomous driving \cite{yu_bdd100k_2020}. Consequently, being trained on domain-specific datasets, they struggle in alternative contexts. As illustrated in Fig.~\ref{fig:system_overview}, a social robot viewing the world from an egocentric, close-quarter perspective faces completely different challenges: humans moving in erratic nonlinear patterns, obstructing each other or walking out of frame \cite{taylor_regroup_2022}. These factors lead to ``pathologically'' missing data that makes standard linear trajectory prediction no longer sufficient \cite{stoler_t2fpv_2023}.

While the literature demonstrates that both face and body tracking offer distinct advantages \cite{wang_real-time_2019, khalifa_face_2022, brown_face_2021}, it remains unclear which modality performs better overall in unstructured HRI. To resolve this, we introduce a focused egocentric dataset to systematically evaluate both modalities under varying configurations of extended memory and appearance re-identification (ReID). While researchers increasingly combine deep learning modules to push tracking robustness \cite{cho_bot-facesort_2025}, our empirical data challenge the assumption that adding more complex tracking features universally improves results in HRI. Specifically, our data suggests that while appearance ReID improves body tracking, applying it to faces destabilizes the tracker, causing IDSW to spike.

This paper provides the following contributions:
\begin{enumerate}
    \item A focused, custom-annotated egocentric HRI dataset, capturing close-range social dynamics not well represented in standard tracking benchmarks.
    \item A systematic evaluation comparing face versus body tracking that isolates tracking from detection errors and assesses the impact of extended memory and appearance ReID.
    \item An optimized pipeline for stationary HRI that reduces IDSW by 49\% against the baseline.
\end{enumerate}

\section{BACKGROUND AND RELATED WORK}

\subsection{Multi-Object Tracking in Computer Vision}
\label{sec:mot_background}
Modern spatial tracking relies on the ``Tracking-by-Detection'' paradigm \cite{bewley_simple_2016}, separating perception into two sequential steps. First, a detector isolates targets without temporal memory. A tracker then links these across frames, typically using a Kalman filter for spatial prediction and the Hungarian algorithm for matching.

Architectures like ByteTrack \cite{zhang_bytetrack_2022} revolutionized this by linking even low-confidence bounding boxes to maintain identities during partial occlusions using Intersection over Union (IoU) spatial overlap. It achieves this via a two-stage matching strategy: associating high-confidence detections first, then performing a second pass to link remaining low-confidence detections with unmatched tracks.

Subsequent models such as BoT-SORT-ReID \cite{aharon_bot-sort_2022} integrate ReID by combining IoU-based spatial costs with cosine distances between visual appearance embeddings extracted from detected person crops. It gates candidate matches using both an appearance threshold and an IoU-based proximity threshold, then forms the final association cost as the minimum of the IoU distance and the  gated cosine distance. This fusion allows appearance information to support associations when spatial overlap is unreliable. Furthermore, to handle moving cameras, the BoT-SORT framework employs Camera Motion Compensation (CMC) using background keypoints. However, the authors note that in scenes with many dynamic objects, this estimation can fail due to a lack of stable background points, causing unexpected tracker behavior.

\subsection{Egocentric Tracking in HRI}

Historically, social robotics relied heavily on RGB-D sensors to capture spatial data \cite{wang_rgb-d-based_2018}. Furthermore, researchers often integrated external sensor networks to overcome the inherently restricted egocentric field of view \cite{han_multiple_2022}. Several systems fused robot vision with static third-person surveillance cameras \cite{lin_joint_2023}, or utilized complementary top-down views to link global trajectories through identity association \cite{han_multiple_2022}. However, to avoid the hardware overhead and constrained applicability of specialized sensors, modern HRI perception has increasingly shifted toward robust, purely monocular 2D Convolutional Neural Network feature extractors \cite{mohsen_real-time_2025}. To maintain identities using only this single viewpoint, architectures designed for autonomous navigation in dynamic environments rely on ``motion-appearance bimodal association,'' demonstrating that appearance features must be integrated with temporal memory to recover targets after dense occlusions \cite{su_q-tracking_2025}.

Facial ReID is crucial for social robots to tailor interactions \cite{wang_real-time_2019}, but applying standard appearance embeddings to faces often leads to tracking degradation. To address this, recent state-of-the-art architectures like BoT-FACE-SORT \cite{cho_bot-facesort_2025} have adapted the BoT-SORT framework by integrating specifically tailored face appearance embeddings. Complementing these facial features, body tracks provide a stable visual anchor. Because bodies capture macroscopic details like clothing \cite{brown_face_2021}, they successfully reduce IDSW during prolonged occlusions when faces are hidden \cite{khalifa_face_2022}. Yet, body tracking alone is insufficient. Environmental obstacles frequently occlude bodies, necessitating face tracking to maintain the interaction's spatial anchor \cite{martin-martin_jrdb_2023, wang_real-time_2019} and provide a biometric corrective layer to resolve identity mix-ups when bodies overlap during close-proximity scenarios \cite{khalifa_face_2022}.

\subsection{Existing Datasets vs. The HRI Reality}

Processing social dynamics strictly through egocentric frames without third-person oracles remains a bleeding-edge goal \cite{scofano_following_2025}. However, evaluating these pipelines is inherently difficult because standard benchmarks \cite{dendorfer_motchallenge_2021} lack robot-mounted, stationary egocentric perspectives \cite{martin-martin_jrdb_2023, ye_tpt-bench_2025}. Large-scale benchmarks like MOT17 and MOT20 are primarily optimized for navigation and surveillance; their focus on linear pedestrian trajectories makes them unsuitable for natural HRI \cite{dendorfer_mot20_2020, dendorfer_motchallenge_2021}. Other datasets, such as DanceTrack \cite{sun_dancetrack_2022}, feature nonlinear motion (NLM) but lack specific conversational transitions. Similarly, datasets like CrowdHuman \cite{shao_crowdhuman_2018} consist only of static images rather than continuous video sequences. Even egocentric datasets like JRDB \cite{martin-martin_jrdb_2023} focus on mobile robots navigating environments rather than stationary social robots engaged in close-quarters conversations. Because egocentric views naturally cause frequent target disappearance \cite{ye_tpt-bench_2025}, we constructed a custom dataset to capture these unstructured, stationary HRI dynamics natively.

\section{FURHAT EGOCENTRIC DATASET}
\label{sec:categories}
\begin{table*}[t]
\vspace*{2mm}   %
\caption{Detailed Composition of the Egocentric HRI Dataset}
\label{tab:dataset_detailed_updated}
\begin{center}
\renewcommand{\tabularxcolumn}[1]{>{\raggedright\arraybackslash}m{#1}}

\begin{tabularx}{\textwidth}{|c|>{\raggedright\arraybackslash}m{1.7cm}||c|X|X|c|c|c|}
\hline
\textbf{Actors} & \textbf{Category} & \textbf{Example} & \textbf{Key Interaction Challenges} & \textbf{Environmental Details}\textsuperscript{1} & \textbf{B}\textsuperscript{2} & \textbf{N}\textsuperscript{3} & \textbf{Length (s)}\\
\hline
\hline
\multirow{4}{*}[-2.8cm]{1} & Basic & \includegraphics[width=2.2cm, valign=c]{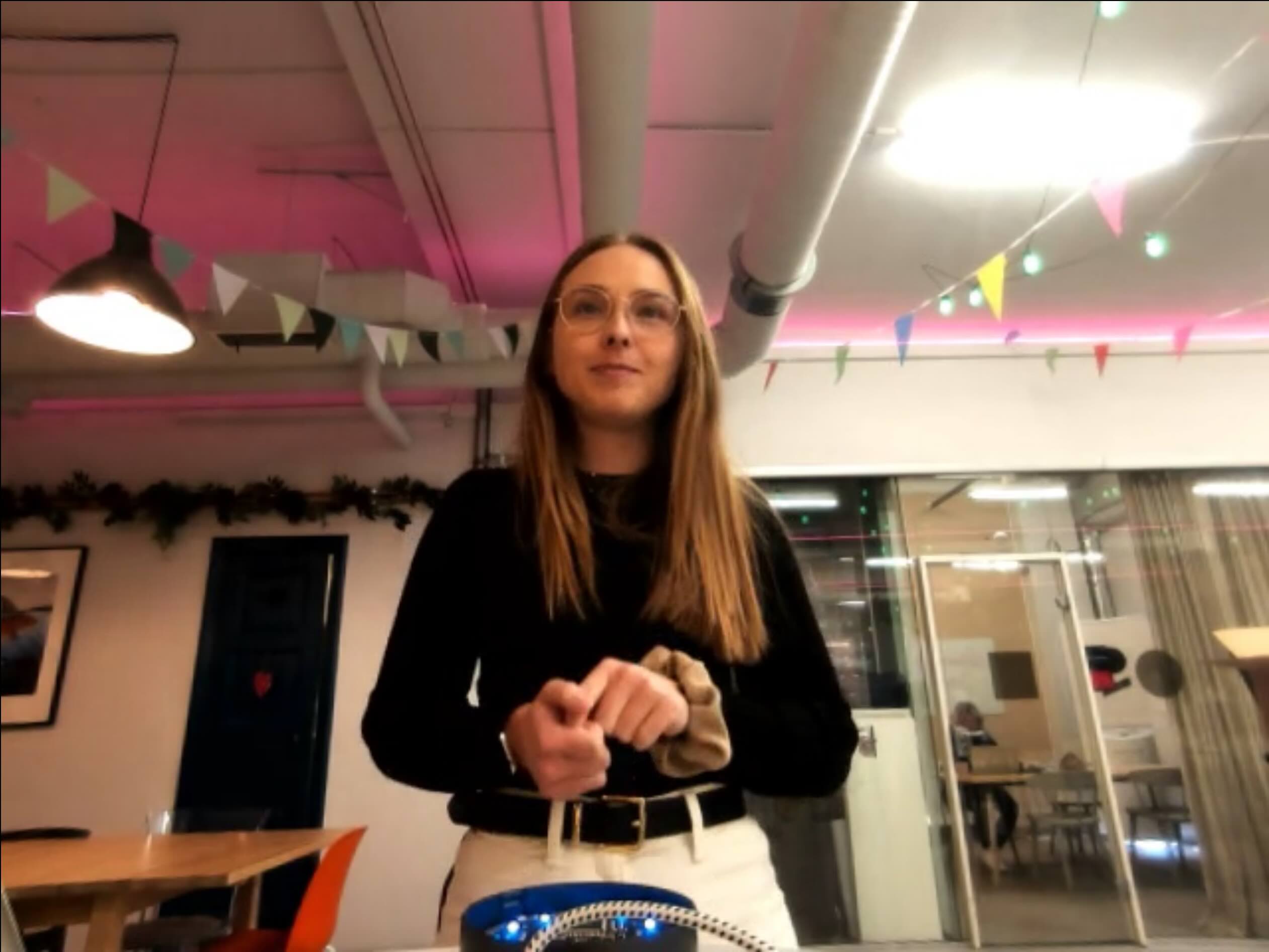} & Main Actor walking in/out from side, standing/``bouncing'' from one foot to another, sitting, stretching, eating nuts. & Body visible from head to belly/hip (Robot Camera at Belly Height). Facing robot frontal. Clean/cluttered background. No occlusion. Dark/Good/Very Good lighting. & 0 & 3 & 150 \\
\cline{2-8}
 & Face \newline Occlusions & \includegraphics[width=2.2cm, valign=c]{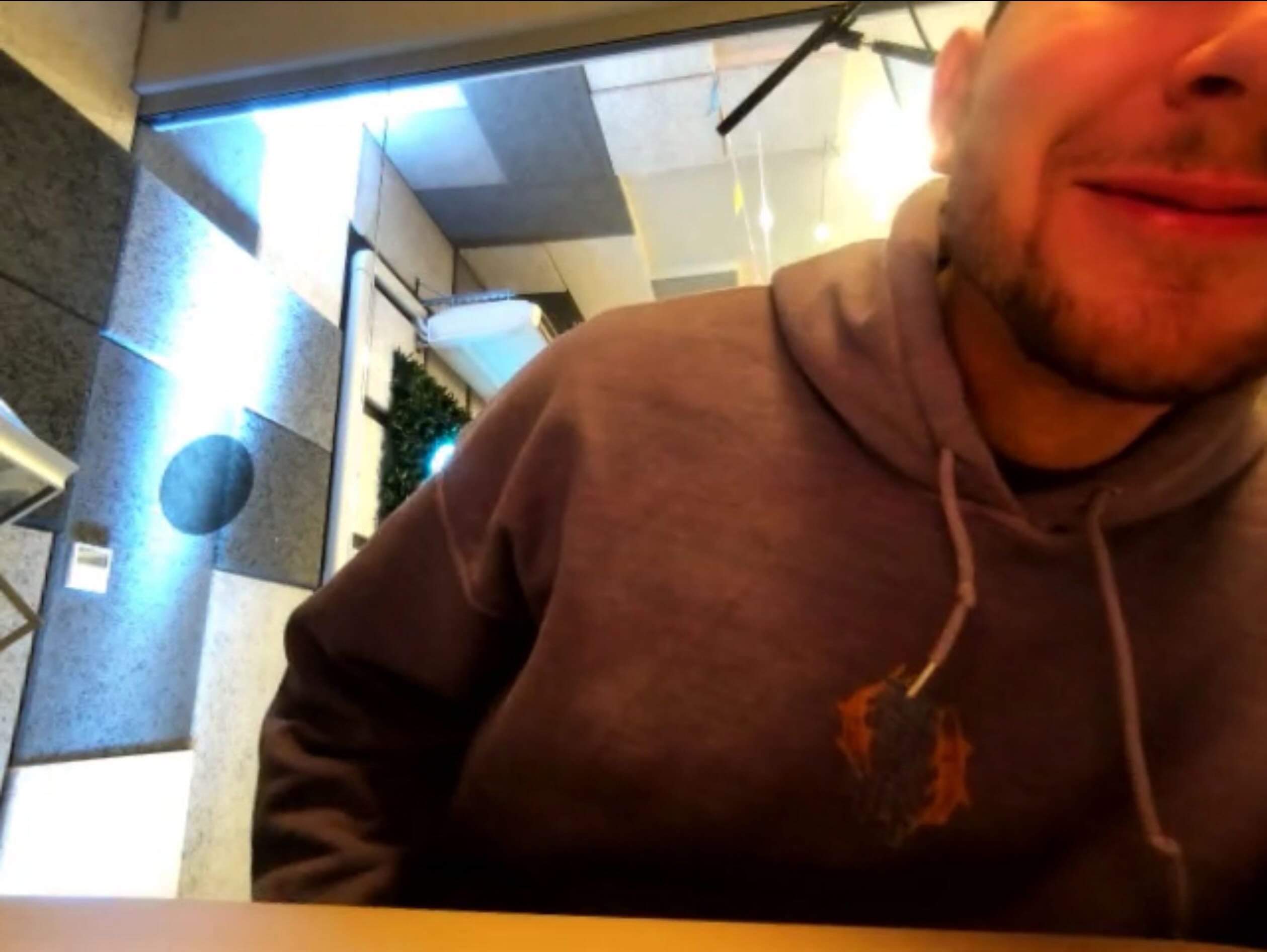} & Main Actor sitting / walking towards robot, moving close with rapid head movements (left/right). Half-face temporarily out-of-screen. Hiding face behind phone/hands/laptops (peeking occasionally). & Body visible from head to chest/knees (Robot Camera at Chest Height). Facing robot frontal. Slightly/very cluttered background. No occlusions by other people. Good lighting. & 0--1 & 4 & 81 \\
\cline{2-8}
 & Not \newline Engaged & \includegraphics[width=2.2cm, valign=c]{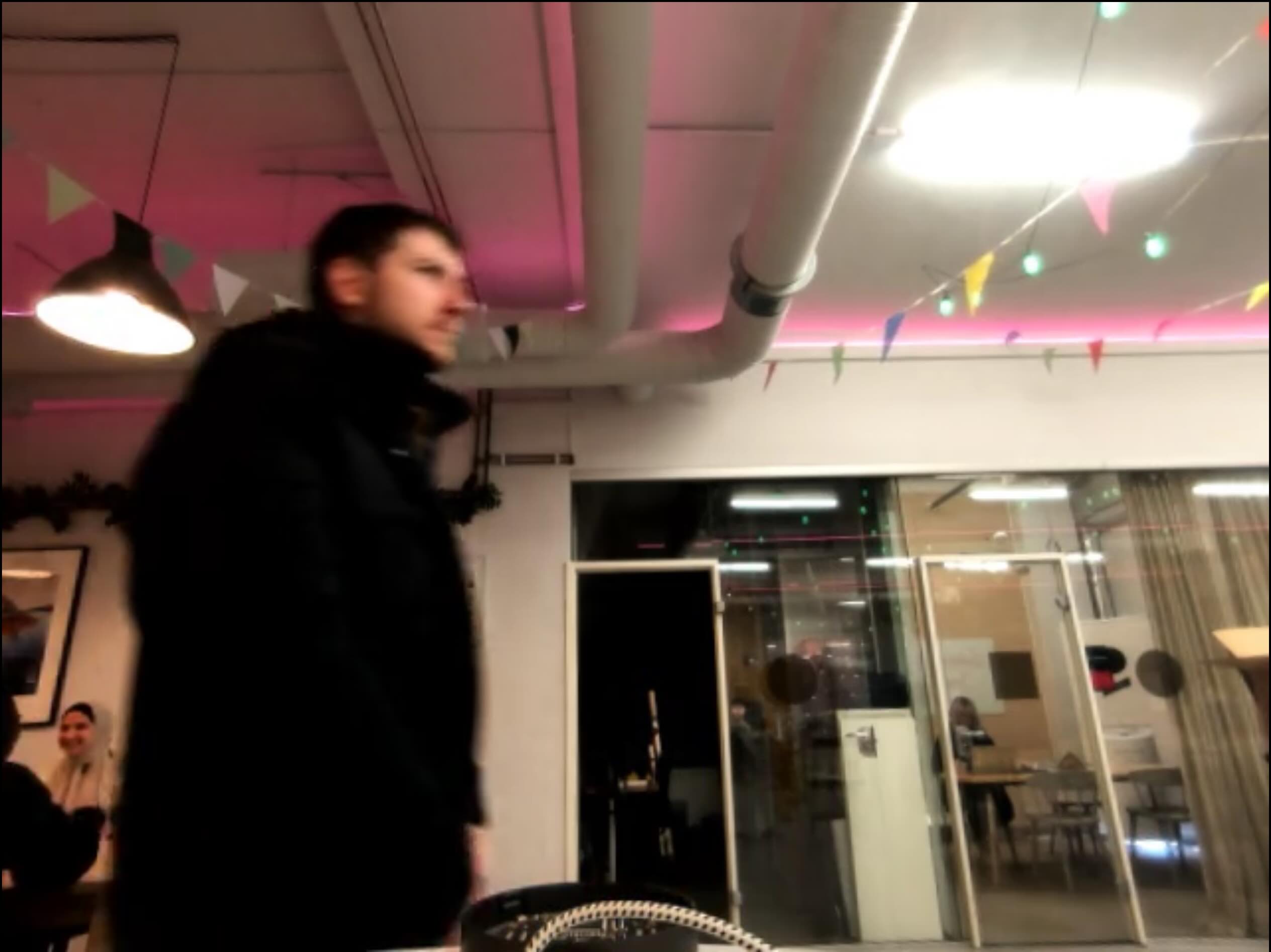} & People walking past the robot/not paying attention. Main Actor walking in/out from side or walking away. Up to 3 bystanders sitting at the table or behind glass wall in the background. & Body visible from head to hip/feet (Robot Camera at Belly Height). Facing robot sideways or from the back. Cluttered background. Occlusions (actor walks in front of bystanders). Good lighting. & 1--3 & 3 & 17 \\
\cline{2-8}
 & Dynamic \newline Background & \includegraphics[width=2.2cm, valign=c]{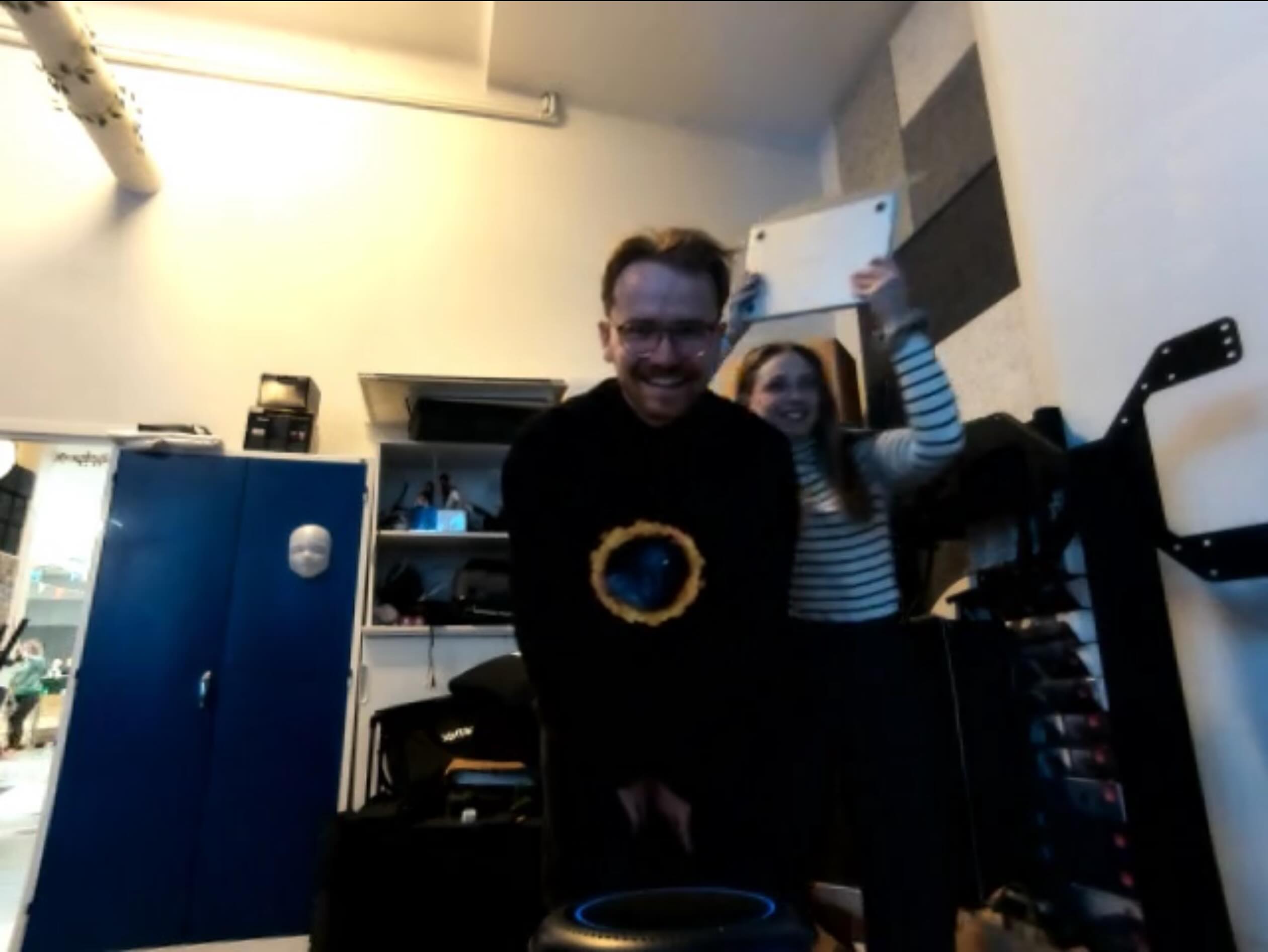} & Main Actor sitting/standing, chatting to robot, occasionally kneeling or covering face with hands. 3--6 bystanders (working, walking past, visible through doors, or actively listening to the conversation). & Body visible from head to chest/knees (Robot Camera at Chest/Hip Height). Facing robot frontal. Cluttered background. Occlusions (bystanders walk in front of/behind main actor). Bad/Good lighting. & 3--6 & 3 & 260 \\
\hline
\multirow{3}{*}[-1.9cm]{2} & Basic & \includegraphics[width=2.2cm, valign=c]{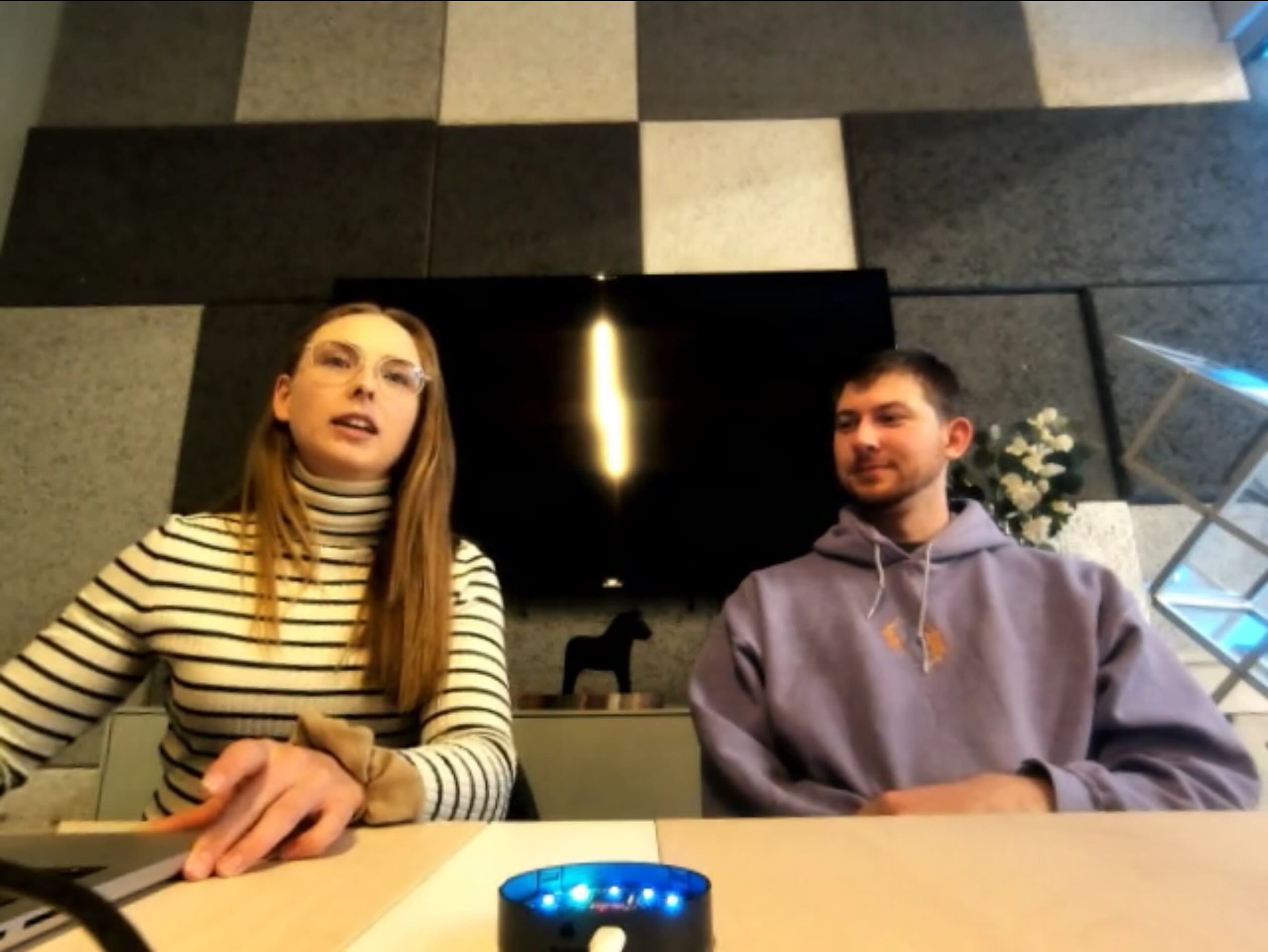} & Main Actors (standing/sitting) engaged in conversation with robot. Social gaze (temporarily turning heads towards each other). One actor temporarily covering face with phone. & Bodies visible from head to chest/hip (Robot Camera at Chest/Belly Height). Facing robot frontal. Clean/cluttered background (with static images of people). No occlusion. OK/Very Good lighting. & 0 & 3 & 596 \\
\cline{2-8}
 & Dynamic \newline Background & \includegraphics[width=2.2cm, valign=c]{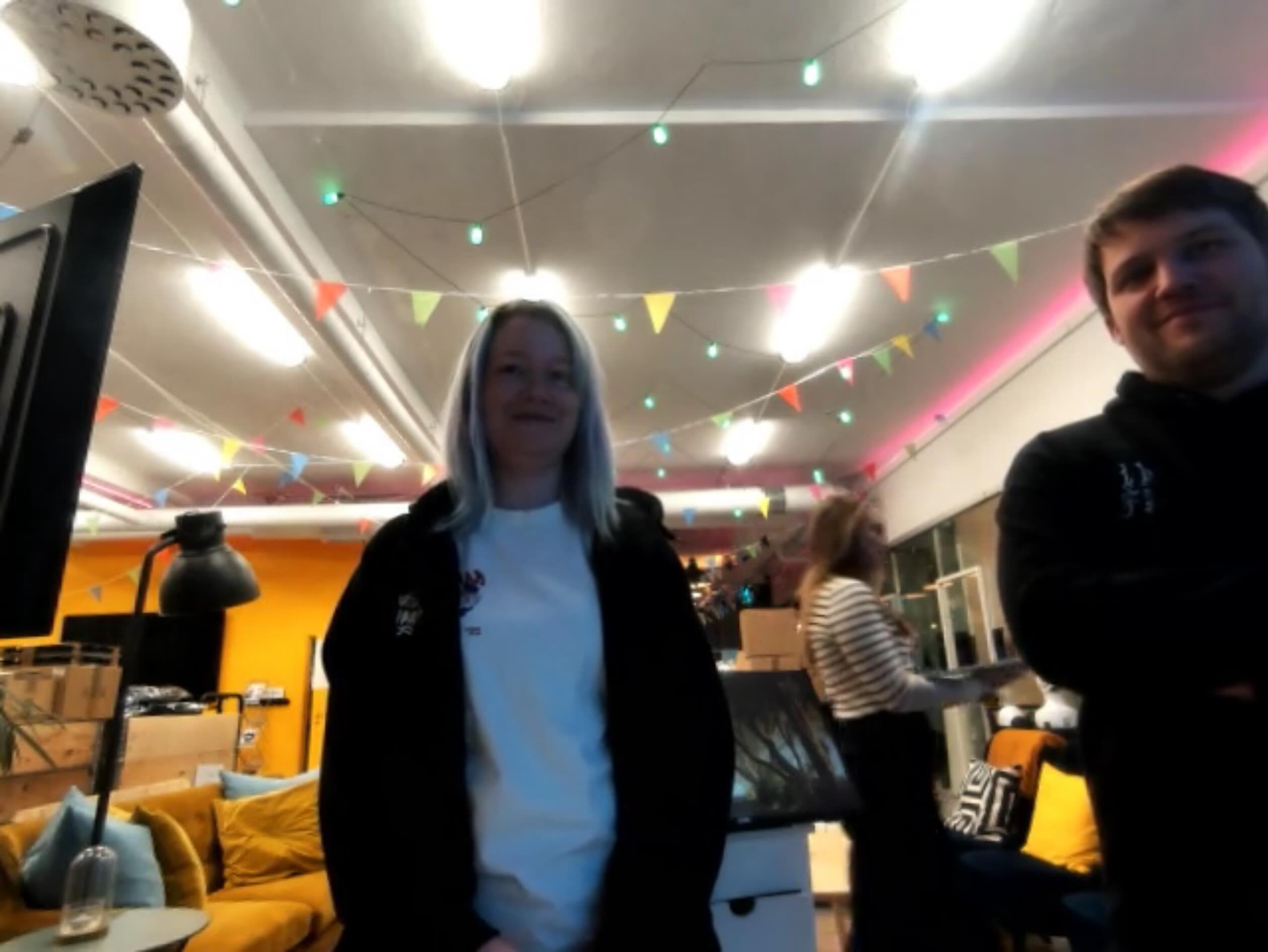} & Main Actors chatting to robot, turning towards each other. Bending over, covering face, turning heads, or halfway in-out-screen. Distant people visible; 1 active bystander walking behind actors. & Bodies visible from head to hip/knee (Robot Camera at Belly/Hip Height). Facing robot frontal. Occasional occlusions (bystander occluded by main actors). Bad/OK lighting. & 0--1 & 2 & 378 \\
\cline{2-8}
 & Crowded & \includegraphics[width=2.2cm, valign=c]{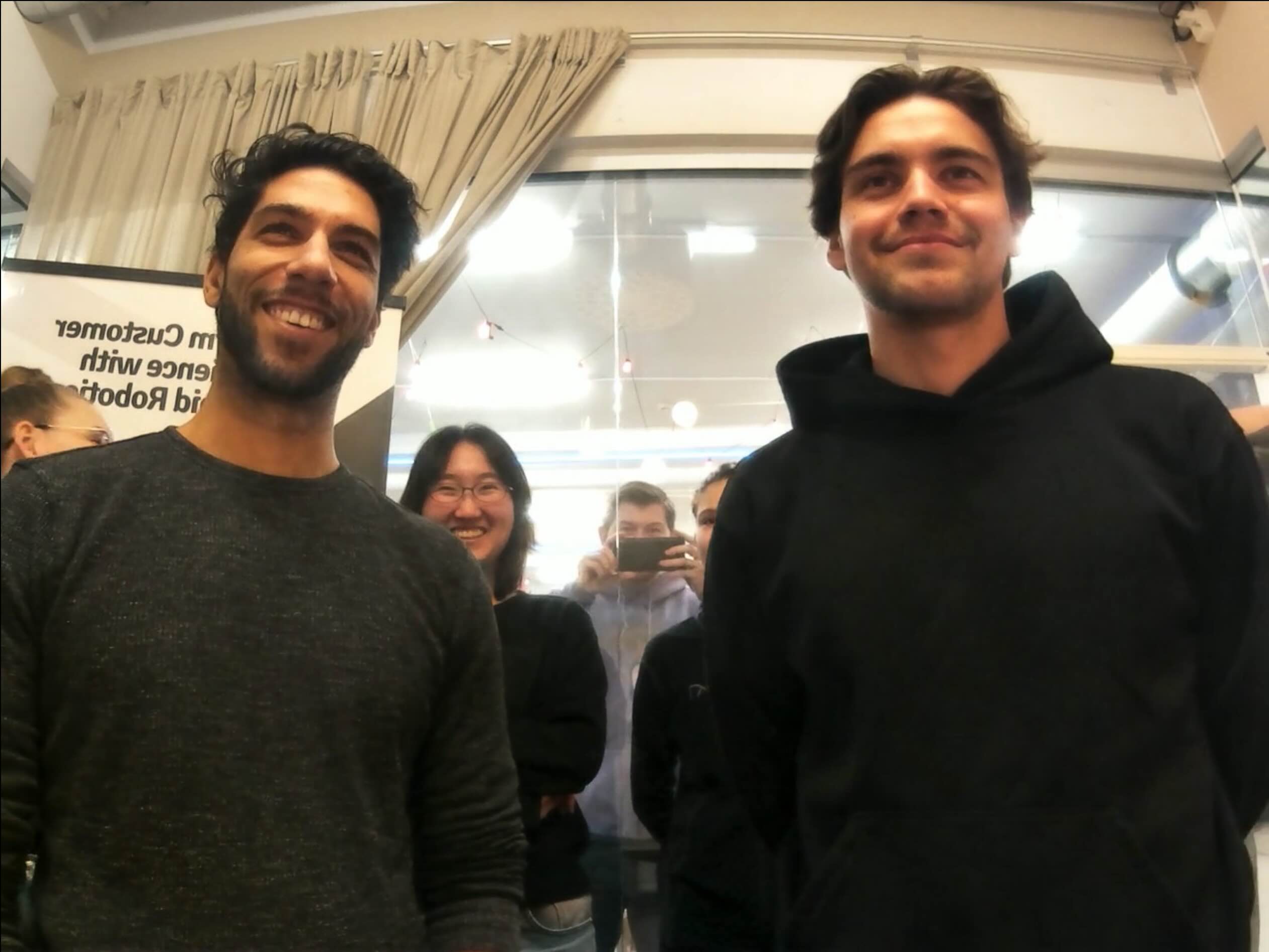} & Main Actors chatting to robot, hand gesticulations, bending, looking back/up. Complex scene with 4--5 very active bystanders (switching positions, peeking over shoulder, using phones). & Bodies visible from head to belly/hip (Robot Camera at Chest/Belly Height). Facing robot frontal. Heavy dynamic occlusions. OK/Good lighting. & 4--5 & 2 & 332 \\
\hline
\end{tabularx}

\vspace{1ex}
{\raggedright \footnotesize 
\textsuperscript{1}All spatial descriptors are defined from the perspective of the main actor; 
\textsuperscript{2}Bystanders per sequence; 
\textsuperscript{3}Number of sequences per category.\par}

\end{center}
\end{table*}

We designed the Furhat egocentric dataset as a focused benchmark for stationary, close-range HRI tracking. It consists of 20 sequences and was collected using the Furhat robot (running \mbox{FurhatOS v2.8.4)}, as seen in Fig.~\ref{fig:system_overview}. We modified the robot's camera core to enable the direct recording of raw video streams and to increase the frame rate from 10 to approximately 25 fps. The Furhat robot is designed specifically for multi-party, face-to-face conversations, so it naturally encourages the kind of dynamic, close-quarters social interaction that our tracking pipeline is intended to address \cite{al_moubayed_furhat_2012}. We capture these scenarios from a stationary, egocentric perspective via the camera integrated into the robot's fixed stand.

The combination of a stationary observer and an unstructured social environment introduces several tracking challenges:
\begin{itemize}
    \item \textbf{Prolonged Occlusions:} Target actors are frequently hidden behind other people or objects for extended periods ($>$ 30 frames).
    \item \textbf{In-Out-Screen Transitions:} People leave the camera's field of view and re-enter later.
    \item \textbf{Nonlinear Motion:} Actors move in erratic patterns that break the linear Kalman filter predictions.
    \item \textbf{U-Turns:} A person disappears behind an occlusion, reverses their direction of motion while hidden and reappears on the unexpected side again.
    \item \textbf{ID Takeover:} A foreground actor walking past a background actor and ``stealing'' their identity.
    \item \textbf{Fragmentation:} Bounding boxes are occasionally dropped for single frames. This occurs frequently at the edges of the video, likely due to fisheye lens distortion.
\end{itemize}

Table~\ref{tab:dataset_detailed_updated} details the composition of the dataset, including 16 unique participants. It is structurally divided by the number of interlocutors (1 or 2) and the specific category of interaction (e.g., \textit{1-Basic}). All videos were recorded at about 25 frames per second with a resolution of 640$\times$480, with the exception of one sequence in the Crowded category, recorded at 1280$\times$960. To establish the ground truth (GT) for our tracking evaluation, we annotated body and face bounding boxes across all frames (45,169) using the Computer Vision Annotation Tool (CVAT) \cite{cvatai_corporation_computer_2024}. Initial proposals were generated with YOLOX (version 0.1.0) \cite{ge_yolox_2021} and RetinaFace (commit \texttt{b984b4b}) \cite{deng_retinaface_2020}, then hand-corrected into identity-consistent amodal bounding boxes. We followed Caltech Pedestrian guidelines \cite{dollar_pedestrian_2012} for bodies and WIDER FACE standards \cite{yang_wider_2016} for faces. The annotations were produced by one annotator only.

\section{EXPERIMENTAL SETUP}

\subsection{Evaluated Configurations and Experimental Progression}
We conducted 11 experiments, in which we systematically analyzed the impact of four architectural components:

\begin{itemize}
    \item \textbf{Tracking Target:} \texttt{B} (Body) or \texttt{F} (Face).
    \item \textbf{Detector:} \texttt{YX} (YOLOX) for Bodies, \texttt{RF} (RetinaFace) for Faces, or \texttt{GT} (Ground Truth).
    \item \textbf{Tracker:} \texttt{BT} (ByteTrack, spatial only) or \texttt{BS} (BoT-SORT, with ReID).
    \item \textbf{Buffer:} Tracker's temporal memory buffer size in frames (\texttt{30}, \texttt{120}, or \texttt{2500}).
\end{itemize}

For the configurations, we adopt the naming convention \texttt{[Target]-[Detector]-[Tracker]-[Buffer]}. To isolate tracking logic from hardware compute constraints, all evaluations were conducted offline following a step-by-step progression: 

\begin{enumerate}
    \item \textbf{End-to-End Baselines:} To establish stable real-world baselines, we chose YOLOX, ByteTrack's default detector \cite{ge_yolox_2021}, for bodies (\textbf{B-YX-BT-30}) and RetinaFace, excelling under severe occlusions and extreme profile angles \cite{deng_retinaface_2020}, for faces (\textbf{F-RF-BT-30}).
    
    \item \textbf{Isolating Tracking Performance:} We then evaluated their GT counterparts (\textbf{B-GT-BT-30}, \textbf{F-GT-BT-30}) to explicitly separate tracking performance from detection errors.
    
    \item \textbf{Extending Temporal Memory:} To address IDSW caused by prolonged occlusions, we evaluated the impact of extended memory on the GT configurations by increasing the buffer size to 120 and 2500 frames (\textbf{B-GT-BT-120}, \textbf{F-GT-BT-120}, \mbox{\textbf{B-GT-BT-2500}}, \textbf{F-GT-BT-2500}).
    
    \item \textbf{Integrating Appearance Features:} Subsequently, we introduced appearance features via BoT-SORT-ReID and BoT-FACE-SORT (\textbf{B-GT-BS-2500}, \textbf{F-GT-BS-2500}) to resolve tracking failures caused by complex NLMs.

     \item \textbf{Real-World Validation:} The most successful tracking configuration was paired back with the standard YOLOX detector (\textbf{B-YX-BS-2500}) to validate the system's end-to-end real-world performance.

    \item \textbf{Environmental Impact Analysis:} Finally, we evaluated the \textbf{B-GT-BT-2500} and \textbf{F-GT-BT-2500} configurations across the interaction categories defined in Table~\ref{tab:dataset_detailed_updated} to isolate specific environmental impacts.
\end{enumerate}

\subsection{Tracker Implementation}
We utilized the official implementation of \textbf{ByteTrack} (base commit \texttt{d1bf019}) \cite{zhang_bytetrack_2022} as our spatial baseline. For appearance-based tracking, we utilized \textbf{BoT-SORT-ReID} (commit \texttt{2519854}) \cite{aharon_bot-sort_2022} for bodies and \textbf{BoT-FACE-SORT} (base commit \texttt{3d597ec}) \cite{cho_bot-facesort_2025} for facial targets. To tailor these ReID architectures to our stationary, egocentric HRI setting, we applied two key modifications:

\begin{itemize}
    \item \textbf{Disabled Camera Motion Compensation:} Because our robot is stationary, we disabled CMC. This prevents moving bystanders from causing unexpected tracker behavior as described in Section~\ref{sec:mot_background}.
    \item \textbf{Disabled Spatial ReID Gating:} We disabled the spatial gating mechanism described in Section~\ref{sec:mot_background} while preserving the underlying cost function \cite{aharon_bot-sort_2022}. This allows a strong visual appearance match to effectively override a poor spatial prediction. This is crucial for successfully linking tracklets, for example during ``U-Turn'' events, where an occluded target reappears in a spatially distant location.
\end{itemize}

\subsection{Evaluation Metrics}
Tracker performance was quantified using the TrackEval framework (commit \texttt{12c8791}) \cite{luiten_hota_2021}. We focused on three primary metrics, interpreted specifically for social robotics:

\begin{itemize}
    \item \textbf{Higher Order Tracking Accuracy (HOTA):} Considered the current ``gold standard'' metric, HOTA combines Detection Accuracy (DetA) and Association Accuracy (AssA) into a single score. It is calculated as the geometric mean of these two components:
\begin{equation}
\mathit{HOTA} = \sqrt{\mathit{DetA} \cdot \mathit{AssA}}.
\end{equation}
    \item \textbf{Identity F1 Score (IDF1):} This metric measures temporal consistency. It is less sensitive to precise bounding box overlap but heavily penalizes identity fragmentation, reflecting how well a user retained the same ID throughout the interaction.
    \item \textbf{IDSW:} The raw count of how many times a single tracked target incorrectly changes its assigned ID. In an HRI context, this is the most critical metric: minimizing IDSW directly correlates to fewer instances where the robot forgets a user and interrupts a natural conversation.
\end{itemize}

\subsection{Qualitative Failure Categorization}
While metrics like HOTA and IDSW indicate that a tracker failed, they do not explain \textit{why}. To bridge this gap, we conducted a manual, qualitative analysis of tracking failures. We conducted a per-sequence review of every IDSW in the GT baseline configurations (\textbf{B-GT-BT-30}, \textbf{F-GT-BT-30}) to map them to the specific HRI challenges established in Section~\ref{sec:categories}. This allowed us to directly map specific social behaviors to architectural tracking weaknesses. Note that due to the high number of IDSW, one sequence from the \textit{Crowded} category was excluded from this manual review.

\begin{table}[b]
\caption{Quantitative Results for Body and Face Tracking}
\label{tab:combined_results}
\begin{center}
\renewcommand{\arraystretch}{1} 
\begin{tabular}{|l||c|c|c|}
\hline
\textbf{Configuration} & \textbf{HOTA} $\uparrow$ & \textbf{IDF1} $\uparrow$ & \textbf{IDSW} $\downarrow$\\
\hline
\hline
\multicolumn{4}{|c|}{\textbf{Body Tracking}} \\
\hline
B-YX-BT-30 (Baseline) & 80.8 & 77.4 & 78\\
\hline
B-GT-BT-30 & 94.8 & 91.1 & 71\\
\hline
B-GT-BT-120 & 96.4 & 95.7 & 59\\
\hline
B-GT-BT-2500 & 96.5 & 96.2 & 54\\
\hline
\textbf{B-GT-BS-2500} & \textbf{98.0} & \textbf{98.1} & \textbf{25}\\
\hline
B-YX-BS-2500 & 89.0 & 90.5 & 40\\
\hline
\multicolumn{4}{|c|}{\textbf{Face Tracking}} \\
\hline
F-RF-BT-30 (Baseline) & 74.3 & 72.2 & 141\\
\hline
F-GT-BT-30 & 88.8 & 86.7 & 128\\
\hline
F-GT-BT-120 & 90.2 & 88.6 & 110\\
\hline
\textbf{F-GT-BT-2500} & \textbf{90.3} & \textbf{88.7} & \textbf{102}\\
\hline
F-GT-BS-2500 & 90.8 & 88.4 & 188\\
\hline
\end{tabular}
\end{center}
\end{table}

\section{RESULTS}
\label{sec:results}

\subsection{Body Tracking Performance}

Table~\ref{tab:combined_results} details body tracking performance. The baseline tracker (B-YX-BT-30) achieved a HOTA score of 80.8\% but suffered from 78 IDSW, indicating frequent identity fragmentation during interactions.

\textbf{Impact of Perfect Detections:} Using GT \mbox{(B-GT-BT-30)} to isolate the tracking logic yielded only a marginal improvement over the baseline, reducing IDSW from 78 to 71. This suggests that many tracking failures are driven by the association algorithm's handling of complex social dynamics rather than by detection inaccuracies.

\begin{figure}[t]
    \centering
    \vspace*{0.7mm}

    \begin{tikzpicture}[trim axis left, trim axis right]
    \begin{axis}[
        width=0.88\columnwidth, %
        height=6cm,            %
        xlabel={Buffer Size (Frames)},
        ylabel={ID Switches},
        xtick={0,1,2},
        enlarge x limits=0.2,
        xticklabels={30, 120, 2500},
        ymin=0, ymax=200,
        grid=major,
        legend columns=2, 
        legend style={
            at={(0.02,0.98)},        
            anchor=north west,      
            font=\footnotesize,        
            inner sep=2pt,          
            column sep=5pt,          
            row sep=-1pt            
        },
        tick label style={font=\footnotesize},
        label style={font=\footnotesize}
    ]
    \addplot[color=blue, mark=*, thick] coordinates { (0,71) (1,59) (2,54) };
    \addlegendentry{Body (Spatial)}

    \addplot[mark=diamond*, mark options={scale=2, fill=cyan}, only marks] coordinates {(2,25)};
    \addlegendentry{Body (ReID)}

    \addplot[color=red, mark=square*, thick] coordinates { (0,128) (1,110) (2,102) };
    \addlegendentry{Face (Spatial)}

    \addplot[mark=x, mark options={scale=2, thick, color=orange}, only marks] coordinates {(2,188)};
    \addlegendentry{Face (ReID)}
    \end{axis}
    \end{tikzpicture}
    
    \caption{Impact of Memory and Appearance on Tracking Stability. Increasing buffer size consistently improves spatial tracking (solid lines slope down). Adding ReID features (at 2500) drastically improves body tracking (diamond drops to 25) but destabilizes face tracking (cross spikes to 188).}
    \label{fig:trends}
\end{figure}
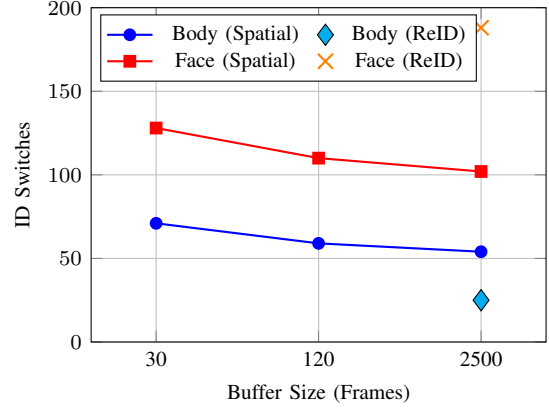

\textbf{Impact of Extended Memory:} Increasing the track buffer to 120 and 2500 frames provided a stepwise stability improvement. The maximum buffer configuration \mbox{(B-GT-BT-2500)} reduced IDSW from 71 to 54 (blue line in Fig.~\ref{fig:trends}), alongside incremental improvements in HOTA and IDF1.

\textbf{Impact of Appearance:} The integration of ReID features (B-GT-BS-2500) provided the final and most substantial stability improvement. This configuration achieved the highest overall scores, reducing IDSW to just \textbf{25} (cyan diamond in Fig.~\ref{fig:trends}), which is a 68\% reduction compared to the baseline.

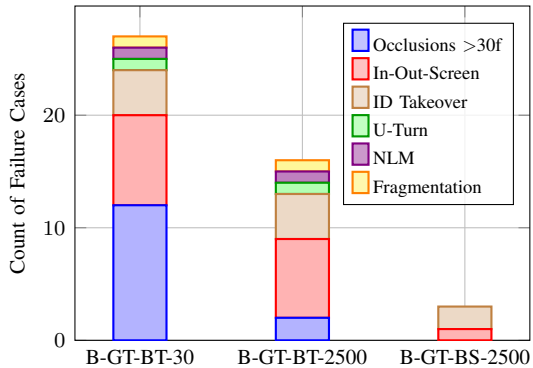
\begin{figure}[b]
    \centering
    \begin{tikzpicture}[trim axis left, trim axis right]
    \begin{axis}[
        width=0.88\columnwidth, %
        height=6cm,            %
        ybar stacked,
        bar width=20pt,
        symbolic x coords={B-GT-BT-30, B-GT-BT-2500, B-GT-BS-2500},
        xtick=data,
        enlarge x limits=0.2,
        ylabel={Count of Failure Cases},
        ymin=0,
        legend cell align=left, 
        legend style={
            at={(0.965,0.95)}, 
            anchor=north east, 
            nodes={scale=0.7, transform shape},
            fill=white, fill opacity=0.8, draw opacity=1, text opacity=1
        },
        grid=major,
        tick label style={font=\footnotesize},
        label style={font=\footnotesize}
    ]
    
    \addplot[ybar, fill=blue!30, draw=blue, thick] plot coordinates {(B-GT-BT-30,12) (B-GT-BT-2500,2) (B-GT-BS-2500,0)};
    \addlegendentry{Occlusions $>$30f}

    \addplot[ybar, fill=red!30, draw=red, thick] plot coordinates {(B-GT-BT-30,8) (B-GT-BT-2500,7) (B-GT-BS-2500,1)};
    \addlegendentry{In-Out-Screen}

    \addplot[ybar, fill=brown!30, draw=brown, thick] plot coordinates {(B-GT-BT-30,4) (B-GT-BT-2500,4) (B-GT-BS-2500,2)};
    \addlegendentry{ID Takeover}

    \addplot[ybar, fill=green!30, draw=green!60!black, thick] plot coordinates {(B-GT-BT-30,1) (B-GT-BT-2500,1) (B-GT-BS-2500,0)};
    \addlegendentry{U-Turn}

    \addplot[ybar, fill=violet!50, draw=violet, thick] plot coordinates {(B-GT-BT-30,1) (B-GT-BT-2500,1) (B-GT-BS-2500,0)};
    \addlegendentry{NLM}

    \addplot[ybar, fill=yellow!50, draw=orange, thick] plot coordinates {(B-GT-BT-30,1) (B-GT-BT-2500,1) (B-GT-BS-2500,0)};
    \addlegendentry{Fragmentation}

    \end{axis}
    \end{tikzpicture}
    
    \vspace{0.5em}
    \caption{Mitigation of qualitative failure modes for body tracking (see Section~\ref{sec:categories} for category definitions). Increasing the buffer size (B-GT-BT-2500) effectively eliminates prolonged occlusions, while the addition of ReID (B-GT-BS-2500) is required to resolve complex IDSW like In-Out-Screen.}
    \label{fig:mitigation}
\end{figure}

\textbf{Failure Mode Analysis:} Fig.~\ref{fig:mitigation} deconstructs these improvements by specific failure types. The extended memory buffer (B-GT-BT-2500) was the primary driver for resolving prolonged ($>$30 frames) but spatially predictable occlusions, reducing them from 12 cases to 2. However, memory alone failed to address more complex events, such as In-Out-Screen transitions and ID Takeovers, for which the ReID module was necessary.

\begin{figure*}[t]
    \centering
    \vspace*{1.8mm}
    \includegraphics[width=0.9\textwidth]{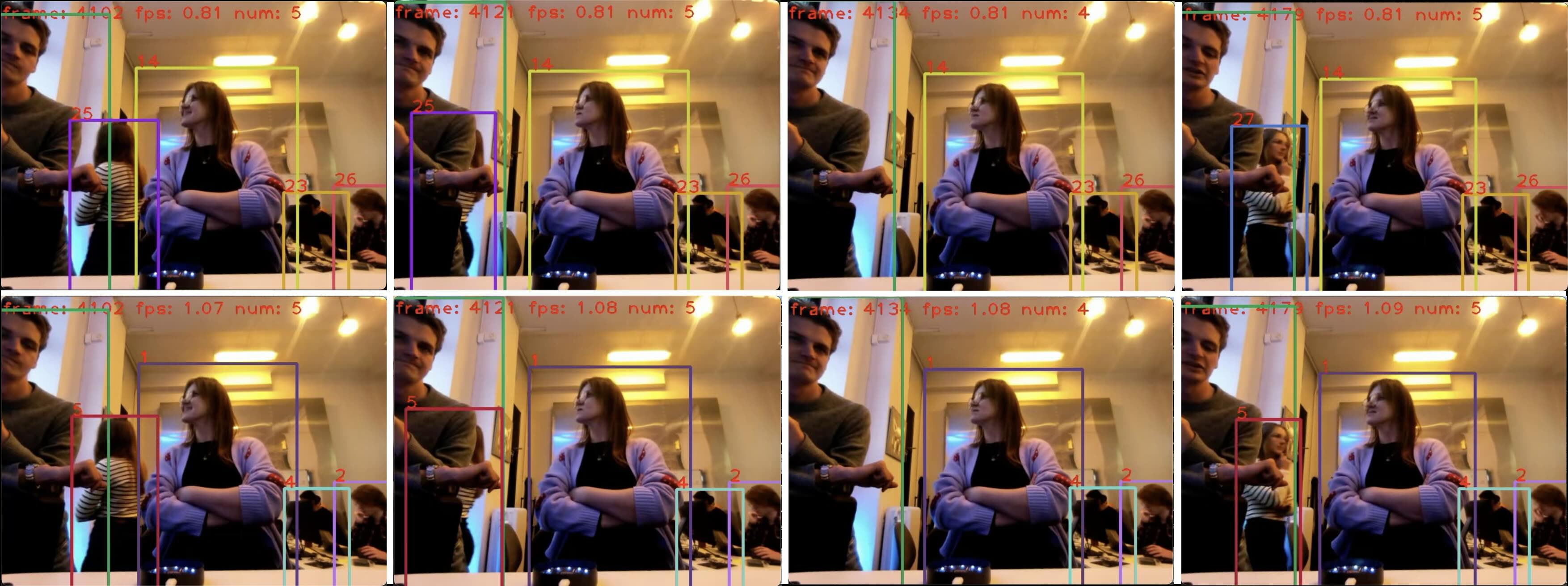}%
    \caption{Qualitative comparison of tracking stability during a complex ``U-Turn'' occlusion event. \textbf{Top Row (Baseline B-YX-BT-30):} The tracker predicts the occluded actor will continue moving left; when the actor turns and reappears on the right, the spatial mismatch causes an IDSW (ID 25 $\to$ ID 27). \textbf{Bottom Row (Proposed B-YX-BS-2500):} Despite the NLM and full occlusion by a second person, the proposed pipeline correctly re-identifies the actor using appearance features, maintaining ID 5 throughout the sequence.}
    \label{fig:qualitative_results}
\end{figure*}

\textbf{End-to-End Real-World Performance:} Finally, we applied this optimized architecture (Extended Buffer + ReID) to the standard YOLOX detector to validate its performance in a real-world setting. Comparing the proposed pipeline \mbox{(B-YX-BS-2500)} against the initial baseline (B-YX-BT-30) shows that the architectural gains transfer effectively: despite using the same detector, the proposed system cut IDSW nearly in half (78 $\to$ 40) and improved the overall tracking accuracy.

\textbf{Resolving Nonlinear Motion:} Fig.~\ref{fig:qualitative_results} visualizes the practical benefit of the proposed body tracking pipeline during a ``U-Turn'' event. In the baseline sequence \mbox{(B-YX-BT-30)}, the tracker loses the actor's ID 25 upon occlusion and re-assigns a new ID 27 when the person reappears on the unexpected side (Top Row). In contrast, the proposed pipeline \mbox{(B-YX-BS-2500)} utilizes ReID to successfully link the tracklets across the occlusion, maintaining ID 5 despite the nonlinear trajectory (Bottom Row).

\subsection{Face Tracking Performance}

Table~\ref{tab:combined_results} details the quantitative performance of the face tracking configurations. The baseline face tracker \mbox{(F-RF-BT-30)} suffered from 141 IDSW, indicating extreme identity fragmentation.

\textbf{Impact of Perfect Detections:} Applying GT bounding boxes for faces (F-GT-BT-30) improved both the baseline HOTA score and the IDSW count (141 $\to$ 128). However, 128 IDSW still represents a high error rate. As with body tracking, this implies that perfect detection accuracy alone may not resolve the complex dynamics of HRI.

\textbf{Impact of Extended Memory:} Increasing the tracker's buffer size (F-GT-BT-2500) further reduced IDSW from 128 to 102. However, comparing the red trajectory (Face) against the blue trajectory (Body) in Fig.~\ref{fig:trends} shows a persistent performance gap: face tracking consistently results in nearly double the error rate across all memory buffers. This implies that faces are harder to track, likely because they are more prone to erratic motion.

\textbf{Failure of ReID:} In contrast to body tracking, enabling ReID caused degradation rather than an improvement. The orange `X' in Fig.~\ref{fig:trends} marks a spike to 188 IDSW, representing an 84\% increase in errors. Our qualitative review of the \mbox{F-GT-BS-2500} tracking errors suggests that the ReID module was highly sensitive to extreme profile angles (looking down, up, or sideways) and dynamic partial occlusions (such as hands or phones temporarily covering the face). Consequently, strict spatial constraints were more effective for maintaining facial identities in this setting.

\subsection{Performance Across Interaction Categories}

To isolate the influence of specific social behaviors and environmental complexities on tracking stability for bodies and faces, we evaluated B-GT-BT-2500 and F-GT-BT-2500 across the interaction categories established in Table~\ref{tab:dataset_detailed_updated}. The results can be found in Table~\ref{tab:category_performance}.

The \textit{Crowded} category accounts for most of the errors (46 IDSW for bodies, 82 for faces). This indicates that navigating dense social scenes with highly active bystanders represents a primary challenge in unstructured HRI environments. Furthermore, as the number of actors increases, body tracking shows greater robustness. Interestingly, face and body tracking perform comparably in \textit{1 - Not Engaged} and \textit{1 - Dynamic Background}. These categories each contain one video with an ``ID Takeover'' event, which we will explore in the following section.

\begin{table}[t]
\caption{Tracking Performance Across HRI Categories (Evaluated using B-GT-BT-2500 for Body and F-GT-BT-2500 for Face)}
\label{tab:category_performance}
\begin{center}
\renewcommand{\arraystretch}{1} 
\begin{tabular}{|c|l|l||c|c|c|}
\hline
\textbf{Actors} & \textbf{Category} & \textbf{Target} & \textbf{HOTA} $\uparrow$ & \textbf{IDF1} $\uparrow$ & \textbf{IDSW} $\downarrow$ \\
\hline
\hline
\multirow{8}{*}{1} & \multirow{2}{*}{Basic} & Body & 99.2 & 100.0 & 0 \\
\cline{3-6}
 &  & Face & 99.1 & 100.0 & 0 \\
\cline{2-6}
 & \multirow{2}{*}{\begin{tabular}[c]{@{}l@{}}Face\\Occlusion\end{tabular}} & Body & 98.6 & 100.0 & 0 \\
\cline{3-6}
 &  & Face & 90.9 & 92.9 & 2 \\
\cline{2-6}
 & \multirow{2}{*}{\begin{tabular}[c]{@{}l@{}}Not\\Engaged\end{tabular}} & Body & 85.5 & 96.3 & 3 \\
\cline{3-6}
 &  & Face & 74.1 & 87.2 & 3 \\
\cline{2-6}
 & \multirow{2}{*}{\begin{tabular}[c]{@{}l@{}}Dynamic\\Background\end{tabular}} & Body & 97.5 & 98.3 & 4 \\
\cline{3-6}
 &  & Face & 91.8 & 89.5 & 5 \\
\hline
\multirow{6}{*}{2} & \multirow{2}{*}{Basic} & Body & 99.8 & 100.0 & 0 \\
\cline{3-6}
 &  & Face & 91.8 & 89.5 & 5 \\
\cline{2-6}
 & \multirow{2}{*}{\begin{tabular}[c]{@{}l@{}}Dynamic\\Background\end{tabular}} & Body & 99.1 & 99.5 & 1 \\
\cline{3-6}
 &  & Face & 85.7 & 83.5 & 9 \\
\cline{2-6}
 & \multirow{2}{*}{Crowded} & Body & 91.5 & 89.7 & 46 \\
\cline{3-6}
 &  & Face & 82.0 & 79.2 & 82 \\
\hline
\end{tabular}
\end{center}
\end{table}

\begin{figure}[t]
    \centering
    \vspace*{1.8mm}
    \includegraphics[width=\columnwidth]{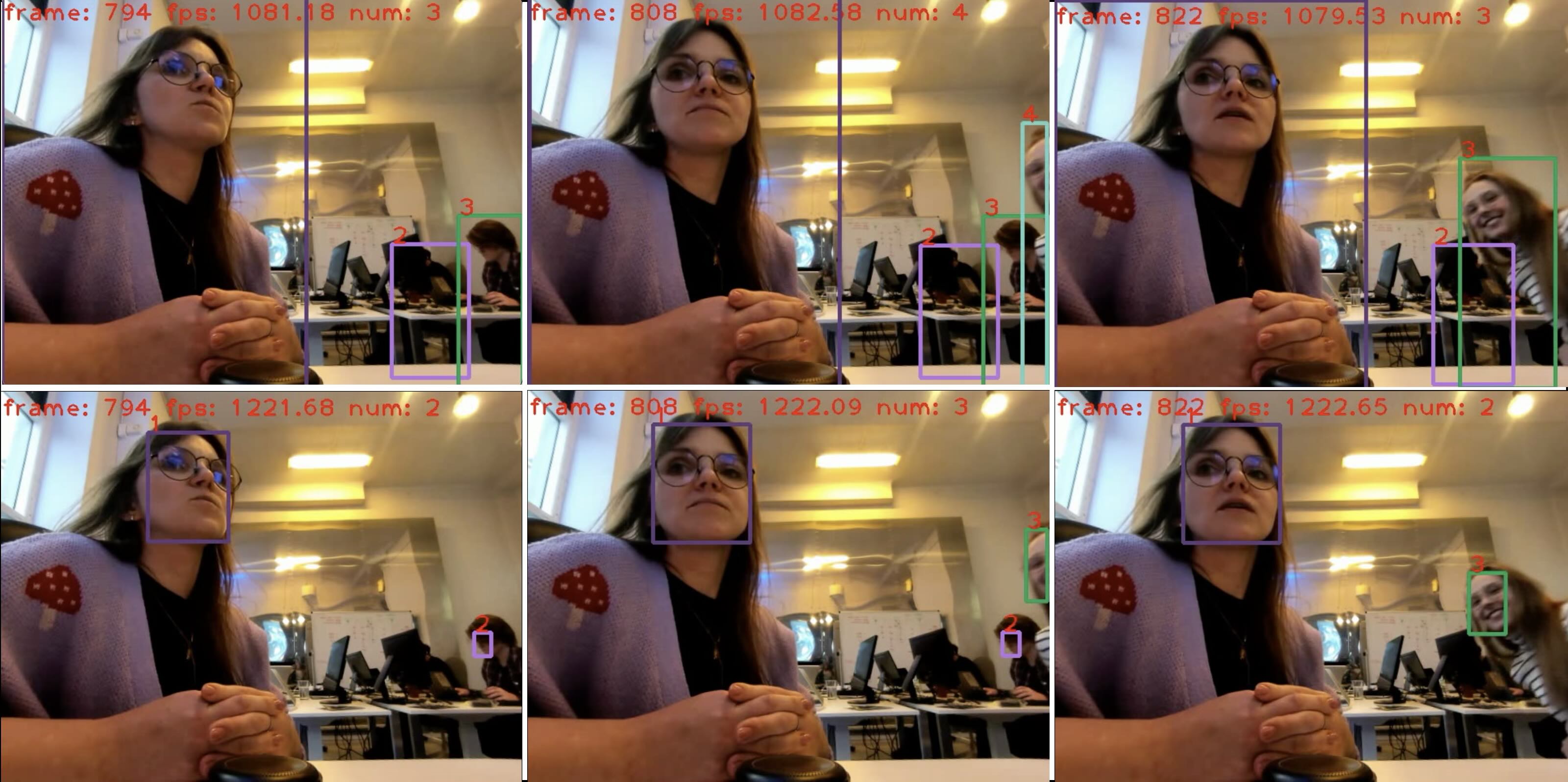}
    \caption{Tracking during an ``ID Takeover'' event. \textbf{Top Row (B-GT-BT-30):} Moving Bystander occludes seated overhearer at the back. The heavy body bounding box overlap causes the Kalman filter to swap their identities. \textbf{Bottom Row (F-GT-BT-30):} Smaller, spatially distinct face boxes minimize IoU overlap, allowing the tracker to correctly maintain individual identities through the dynamic occlusion.}
    \label{fig:id_takeover}
\end{figure}

\subsection{Edge Case Analysis}

While body tracking produced better global metrics, our qualitative analysis revealed a specific scenario where face tracking was more robust: ID Takeovers. When actors cross paths, their body bounding boxes overlap heavily. This high IoU frequently confuses the spatial association algorithm. For example, in a \textit{1-Dynamic-Background} sequence, one bystander temporarily occludes another in the background. The resulting spatial ambiguity caused the body tracker (B-GT-BT-30) to swap their identities. In contrast, facial bounding boxes are smaller and more localized. Even when bodies significantly overlap, faces typically remain spatially distinct. Consequently, as illustrated in Fig.~\ref{fig:id_takeover}, the face tracker (F-GT-BT-30) successfully maintained the correct identities during these  crossing events.

\section{DISCUSSION}
Our findings suggest that while robust spatial detection is important, the primary bottleneck for HRI tracking remains temporal association. To address this, increasing the memory buffer proved highly effective at mitigating prolonged occlusions. However, there is a significant performance gap between modalities: face tracking consistently results in higher error rates across all buffers compared to body tracking, showing that faces are harder to track, likely because they are more prone to erratic motion. Nevertheless, regardless of modality, memory alone failed to address more complex dynamic events.

To resolve these more complex IDSW, such as In-Out-Screen transitions and ID Takeovers, the ReID module was necessary. However, it exhibited opposing effects depending on the target. Integrating visual appearance features into body tracks substantially improved overall tracking accuracy and mitigated identity fragmentation. The body likely acts as a stable visual anchor that remains consistent even during erratic movement. Conversely, applying ReID to faces caused IDSW to spike, as the module was highly sensitive to extreme profile angles and dynamic partial occlusions. Therefore, standard facial embeddings appear less suited for close-quarters HRI.

Ultimately, neither body nor face tracking is sufficient alone. While body tracking provides excellent global stability, it fails during heavy bounding box overlap, causing \mbox{``ID Takeovers''}. In contrast, face tracking excels in these exact crossing events because facial bounding boxes are smaller and typically remain spatially distinct. Furthermore, face tracking is essential for downstream feature extraction and provides the exact spatial anchor a robot requires to direct its gaze and maintain eye contact. HRI tracking pipelines may therefore benefit from fusing face and body tracking.

\section{LIMITATIONS AND FUTURE WORK}

While our pipeline demonstrates substantial stability improvements, the generalizability of our findings is constrained by our experimental design. First, our dataset comprises only 20 sequences captured from a stationary viewpoint in an indoor office environment with a small number of participants and limited environmental diversity. Second, evaluating solely close-quarter, face-to-face interactions leaves distant or nonconversational encounters unassessed. Third, our reliance on the Furhat robot's specific wide-angle fisheye lens means observations regarding peripheral trajectory fragmentation may not generalize to robots utilizing different cameras. Finally, this offline, vision-only evaluation does not account for real-time compute constraints, complementary audio cues from visually occluded or out-of-frame participants, or the variability of ReID performance across models.

Positioning this work as an initial step, future work should expand dataset scale and environmental diversity. A logical next step involves transitioning this pipeline to live deployment on the Furhat robot to evaluate real-world latency and its impact on interaction quality. Finally, further investigation into the dynamic fusion of face and body tracking shows promise in resolving complex occlusions more effectively in unstructured environments.

\section{CONCLUSION}

This study indicates that a key challenge of egocentric HRI tracking lies in temporal association rather than spatial detection alone. Our systematic evaluation highlights distinct advantages for different tracking modalities: body tracking paired with appearance ReID excels at maintaining global stability through complex occlusions, while face tracking with strict spatial constraints is superior for resolving identity takeovers during close-proximity crossing events. Ultimately, by optimizing temporal memory and appearance features for the specific dynamics of social interactions, our proposed pipeline (B-YX-BS-2500) successfully reduced IDSW by 49\% compared to the baseline. This performance improvement directly mitigates interaction breakdowns, allowing a social robot to effectively maintain its conversational footing.

\section*{DATA AVAILABILITY}
The dataset is available upon reasonable request to the first author. To protect participants' privacy, access requires a signed Data Usage Agreement.

\section*{ACKNOWLEDGMENT}
Gemini was used for literature search, language editing, \LaTeX\ formatting and data visualization; the authors take full responsibility for the content. Per Swedish law, formal ethical review was not required for the collection of this dataset. Informed consent was obtained for data collection and image publication.

\addtolength{\textheight}{-2.1cm}   %
\bibliographystyle{IEEEtran}
\bibliography{refs}

\end{document}